\documentclass[11pt,a4paper,twocolumn]{article}

\usepackage[top=1in,bottom=1in,left=0.75in,right=0.75in]{geometry}
\usepackage{amsmath,amssymb,amsthm}
\usepackage{graphicx}
\usepackage{booktabs}
\usepackage{xcolor}
\usepackage{hyperref}
\usepackage{natbib}
\usepackage{microtype}
\usepackage{enumitem}
\usepackage{tikz}
\usetikzlibrary{arrows.meta,positioning}
\usepackage{caption}

\newtheorem{definition}{Definition}

\hypersetup{colorlinks=true,linkcolor=black,citecolor=black,urlcolor=black}

\title{\large\textbf{When Does Belief-Based Agent Memory Help?\\
Reliability-Conditional Updating and Provenance-Capped Poisoning Defense}}
\author{
  Pranav Singh \\
  Indian Institute of Technology Ropar \\
  \texttt{2023mcb1308@iitrpr.ac.in}
}
\date{\today}

\begin{document}
\maketitle
\thispagestyle{empty}

\begin{abstract}
We study \emph{when} a belief-based memory actually helps an LLM agent, and find
the answer is conditional. Our vehicle is Nous, a memory that stores knowledge
not as records but as a categorical probability distribution per
entity-attribute pair, updated by a closed-form Bayesian posterior scored by
information-theoretic surprise ($\mathcal{S}=-\log_2 P$), with deltas as the
primary artifact and entropy decay as forgetting. Our findings are analytical,
not a leaderboard claim. First, the Bayesian belief update is \emph{inert} on the
LoCoMo benchmark~\citep{maharana2024locomo}: on a three-conversation subset,
replacing it with naive last-write-wins does not reduce accuracy, because ordinary
conversational QA rarely presents contradictory, low-reliability evidence. (Under
strict token-F1 on the full benchmark, Nous and the strongest academic baseline
A-MEM~\citep{xu2025amem} split the four question types two--two.) Second, the update
pays off only when observations carry per-source \emph{reliability}; we estimate it
from epistemic markers in language, and on a controlled contradiction/staleness
benchmark this recovers a large advantage for belief-updating over last-write-wins
(100 vs 67) and a strong LLM-over-raw-memory baseline (100 vs 90), concentrated in
noise resistance. Third, that reliability signal is an attack surface:
\emph{content}-inferred trust is gameable (a confidently phrased poison earns
reliability $0.96$), so we derive trust from source \emph{provenance}, composed as
$\min(\text{provenance},\text{content})$, unlike prior content-based
defenses~\citep{mempoison2026}. This holds memory-poisoning attack success at $0\%$
under a volumetric flood where naive baselines reach $100\%$; it is bounded, and
requires taint-tracking we characterize but do not implement. Finally, we document
a 27.5-point macro gap between strict token-F1 and a generous LLM-judge on
identical LoCoMo outputs (59.97 vs 32.45), a caution for memory-system comparison.
All code and both benchmarks are public.
\end{abstract}

\section{Introduction}

A persistent limitation of large language model (LLM) agents is the absence
of a continuously evolving memory. The context window provides short-term
recall within a session, but once a session ends all conversational history
is lost unless externally persisted. This has motivated a growing body of
work on agent memory systems, mechanisms that store, retrieve, and reason
over information accumulated across many sessions.

Existing approaches treat memory as a database of facts. Retrieval-augmented
generation (RAG)~\citep{lewis2020rag} encodes conversation chunks as dense
vectors and retrieves the $k$-nearest at query time. Knowledge-graph systems
such as Zep~\citep{zep2024graphiti} store subject-predicate-object triples.
Episodic systems such as MemGPT~\citep{packer2023memgpt} maintain explicit
core memory strings and flush older records to an archival store. More recent
systems including Mem0~\citep{chhikara2025mem0} and A-MEM~\citep{xu2025amem}
combine graph and vector representations to improve retrieval precision.

Despite their diversity, all these systems share one assumption: memory is a
collection of things that happened. This produces three structural failure
modes. First, conflict blindness: when a user changes jobs or preferences,
contradictory records accumulate with no principled resolution mechanism.
Second, heuristic forgetting: deletion rules are time-based rather than
semantic. Third, retrieval granularity mismatch: a question asking what a
person has painted over the years requires aggregating values from many
sessions, but a fact store returns individual records with no accumulation
mechanism.

We argue these failures share a root cause: treating memory as storage rather
than as a model of the world. The brain does not store a verbatim transcript
of past conversations; it maintains a generative model of its environment and
updates it when observations diverge from predictions
\citep{rao1999predictive,friston2010fep}. Surprise drives learning. Unused
beliefs decay toward uncertainty. Memory is a compressed, continuously refined
probability distribution over the state of the world.

Nous (from the Greek word for mind or intellect) implements this view as a
practical agent memory system. The core data structure is a dimension, a
categorical distribution over the possible values of a single entity-attribute
pair. New observations are scored by information-theoretic surprise and
integrated via Bayesian posterior update. The record of that update, a delta,
is the primary stored artifact, capturing not what was observed but how it
changed the agent's understanding. Forgetting is entropy decay, and conflict
resolution is implicit: contradictory evidence naturally shifts probability
mass away from the old value.

This paper is a mechanistic study rather than a leaderboard entry, and its
contributions are three. \textbf{(1) A reliability-conditional account of when
belief-updating helps.} A controlled ablation shows the Bayesian update is inert
on LoCoMo (a three-conversation subset); naive last-write-wins ties it, because
the benchmark rarely presents contradictory, low-reliability evidence; the update
then recovers a large, mechanism-attributable advantage once supplied a
per-observation reliability signal (estimated from epistemic markers in language),
on a contradiction/staleness benchmark where it beats both last-write-wins and a
strong LLM-over-memory baseline. \textbf{(2) Provenance-capped belief updating as a
memory-poisoning defense.} Because the reliability signal is itself an attack
surface, we contribute (narrowly, and \emph{not} as the first trust-based
defense~\citep{mempoison2026}) four results: trust derived from source
\emph{provenance} rather than content; a measurement that content-inferred trust is
adversarially gameable; the $\min(\text{provenance},\text{content})$ composition
rule under which a volumetric poisoning flood fails ($0\%$ vs $100\%$ for naive
baselines); and the taint-tracking requirement without which the defense is
laundered away. \textbf{(3) A quantification of LoCoMo metric fragility}: a
27.5-point macro gap between strict token-F1 and a generous LLM-judge on identical
outputs (59.97 vs 32.45), which we argue should temper leaderboard comparisons.
The architecture (dimensions, deltas, surprise, entropy decay) and a
dependency-free open-source implementation are the vehicle for these findings, not
the claim; we describe them in Section~\ref{sec:arch}.

\section{Background and Related Work}

\paragraph{Long-term conversational memory.}
The LoCoMo benchmark~\citep{maharana2024locomo} evaluates agent memory over
conversations spanning up to 35 sessions, 300 to 600 turns, and 9,000 to
16,000 tokens, covering single-hop, multi-hop, temporal, and open-domain
questions. LongMemEval~\citep{wu2025longmemeval} (ICLR 2025) extends this
with harder, audited questions and is increasingly adopted as a complementary
standard.

\paragraph{RAG-based memory.}
RAG~\citep{lewis2020rag} retrieves the top-$k$ embedding-nearest chunks. Its
primary failure is the inability to aggregate information distributed across
many chunks, compounded by the lost-in-the-middle effect~\citep{liu2023lost}
wherein relevant evidence buried in long contexts is ignored.

\paragraph{MemGPT.}
MemGPT~\citep{packer2023memgpt} partitions memory into in-context, recall,
and archival tiers managed via function calls. It is effective for
single-session recall but lacks conflict-resolution semantics. On LoCoMo with
GPT-4o-mini it achieves token-F1 of 26.65 (single-hop) and 9.15
(temporal)~\citep{xu2025amem}.

\paragraph{A-MEM.}
A-MEM~\citep{xu2025amem} enriches memories with Zettelkasten-style links,
forming a graph of interconnected notes with LLM-generated attributes. It is
the strongest peer-reviewed academic baseline on LoCoMo, reporting token-F1 of
27.02, 45.85, 12.14, and 44.65 on single-hop, multi-hop, temporal, and
open-domain with GPT-4o-mini~\citep{xu2025amem}.

\paragraph{Zep and Mem0.}
Zep~\citep{zep2024graphiti} builds a temporal knowledge graph; Mem0
\citep{chhikara2025mem0} maintains a hybrid vector and graph store. Both
report strong LoCoMo scores but use GPT-4o with a lenient LLM-as-judge.
An independent audit by Penfield Labs~\citep{locomo2025audit} found this judge
accepts up to 63 percent of intentionally wrong answers, and that approximately
6.4 percent of LoCoMo ground-truth answers contain errors.
We therefore restrict our numerical comparison to A-MEM and MemGPT, which report
strict token-F1, and, because this judge fragility is central, we report both
strict token-F1 and an LLM-judge score for Nous and quantify their gap
(Section~\ref{sec:results}) rather than reporting the judge alone.

\paragraph{Predictive coding and the Bayesian brain.}
Nous draws theoretically on predictive coding~\citep{rao1999predictive}, which
models the brain as a generative model minimising prediction error, and on
the Free Energy Principle~\citep{friston2010fep}, which unifies perception and
learning under surprise minimisation. The Bayesian brain
hypothesis~\citep{knill2004bayesian} grounds this computationally. Shannon's
information measure~\citep{shannon1948} supplies the definition
$\mathcal{S}(x) = -\log_2 P(x)$ bits, and Cover and Thomas~\citep{cover2006}
supply the entropy and mutual information foundations.

\paragraph{Concurrent belief-based agent memory.}
A parallel line of recent work also represents memory probabilistically.
BeliefMem~\citep{liao2026beliefmem} stores multiple candidate conclusions as
independent probability-carrying entries, updated by a Noisy-OR rule as evidence
accumulates, and evaluates on LoCoMo and ALFWorld; the Belief
Engine~\citep{yang2026beliefengine} maintains an auditable log-odds stance per
proposition for multi-agent deliberation; and graph-native cognitive
memory~\citep{park2026graphnative} applies symbolic (AGM-style) belief revision
over versioned memory. Nous differs in representation, a single closed-form
\emph{categorical} Bayesian posterior per entity-attribute dimension, rather than
a Noisy-OR over independent hypotheses or a scalar log-odds stance, but the
load-bearing difference is one of question: these systems propose a probabilistic
memory and measure its accuracy, whereas our contributions are the
\emph{reliability-conditional} account of \emph{when} such memory helps
(Section~\ref{sec:reliability}) and the provenance-capped poisoning defense
(Section~\ref{sec:security}), neither of which they address.
BeliefShift~\citep{myakala2026beliefshift}, a benchmark for temporal belief
consistency and opinion drift across sessions, is a natural target for the
real-dialogue evaluation we leave to future work.

\paragraph{Memory poisoning and trust-based defenses.}
A persistent-state attack surface unique to agents is \emph{memory poisoning}:
an attacker writes malicious content into long-term memory so the agent acts on
it in later sessions. AgentPoison~\citep{chen2024agentpoison} optimises a
backdoor trigger in a RAG/memory store (attack success $\ge 80\%$ at under
$0.1\%$ poison rate), and MINJA~\citep{dong2025minja} injects malicious records
through query-only interaction (76.8\% attack success), and sleeper
poisoning~\citep{pulipaka2026sleeper} plants dormant poisoned memories that
activate only in later sessions; these establish that memory poisoning is a
real, studied threat, which OWASP now lists
as ASI06~\citep{owasp2026agentic}. We cite these to motivate the threat, not as
a numerical baseline: their attack shapes (optimised trigger backdoor,
reasoning-bridge injection) differ from the fact-overwrite setting we study, so
success rates are not directly comparable. On defense, the closest prior work
assigns \emph{content-derived} trust: \citet{mempoison2026} score trust from
composite content and pattern signals with trust-aware retrieval and temporal
decay, and A-MemGuard~\citep{wei2025amemguard} detects poisoning by consensus
across an agent's reasoning paths, both content- or pattern-based signals. We
show (Section~\ref{sec:security}) that content-inferred trust is
adversarially gameable (a confident-sounding injection earns high trust) and
that \emph{provenance}-derived trust, composed as
$\min(\text{provenance},\text{content})$ and propagated through trusted
intermediaries via taint tracking, is what actually resists poisoning;
concurrent lineage-tracking memory~\citep{ouyang2026memlineage} builds the
derivation graph such propagation requires. OWASP
already recommends provenance tracking as a mitigation~\citep{owasp2026agentic};
our contribution is not the concept but a concrete mechanism for it and an
empirical map of when it holds, where it fails, and what it costs.

\section{The Nous Architecture}
\label{sec:arch}

\subsection{Core Abstractions}

\begin{definition}[Dimension]
A dimension $\mathcal{D}_{e,a}$ for entity $e$ and attribute $a$ is a pair
$(V,\, p)$ where $V = \{v_1, \ldots, v_n\}$ is a finite vocabulary and
$p : V \to [0,1]$ satisfies $\sum_{v} p(v) = 1$.
\end{definition}

\medskip

A dimension encodes what the agent believes about one aspect of one entity.
For instance, $\mathcal{D}_{\text{Pranav},\,\text{employer}}$ might be
$\{(\text{Google},\, 0.08),\; (\text{Sarvam AI},\, 0.87),\;
(\text{unknown},\, 0.05)\}$. The mode $\hat{v} = \arg\max_v p(v)$ is the
current best belief.

\begin{definition}[Delta]
A delta $\Delta$ is a tuple
$\langle e,\; a,\; p_{\mathrm{prior}},\; p_{\mathrm{post}},\;
\mathcal{S},\; \tau,\; \omega \rangle$,
where $p_{\mathrm{prior}}$ and $p_{\mathrm{post}}$ are the distributions
before and after the update, $\mathcal{S}$ is surprise in bits, $\tau$ is
the timestamp, and $\omega$ is the supporting evidence text.
\end{definition}

\medskip

The delta is the primary stored artifact of Nous. It records not what happened
but how the agent's understanding changed. The full delta sequence for a
dimension is an auditable history of every belief revision.

\subsection{Information-Theoretic Surprise}

Given dimension $\mathcal{D}_{e,a} = (V, p)$ and observed value obs, the
surprise is:

\begin{equation}
  \mathcal{S}(\text{obs} \mid \mathcal{D}_{e,a})
    = -\log_2\, p(\text{obs})
  \quad \text{bits.}
  \label{eq:surprise}
\end{equation}

If obs $\notin V$, the vocabulary is extended with $p(\text{obs}) = \varepsilon$
and masses renormalised, yielding near-maximum surprise for novel values. An
observation confirming the current belief causes negligible change; one that
contradicts it forces a large revision, mirroring the role of prediction error
in biological memory consolidation~\citep{friston2010fep}.

\subsection{Bayesian Update}

Nous computes the posterior via Bayes' rule~\citep{bayes1763}. The likelihood
is:

\begin{equation}
  \mathcal{L}(\text{obs} \mid v) =
  \begin{cases}
    1 - \varepsilon & v = \text{obs}, \\[4pt]
    \dfrac{\varepsilon}{\,|V|-1\,} & \text{otherwise,}
  \end{cases}
  \label{eq:likelihood}
\end{equation}

and the posterior is:

\begin{equation}
  p_{\mathrm{post}}(v) =
  \frac{\mathcal{L}(\text{obs} \mid v)\; p_{\mathrm{prior}}(v)}
       {\displaystyle\sum_{v' \in V}
        \mathcal{L}(\text{obs} \mid v')\; p_{\mathrm{prior}}(v')}.
  \label{eq:bayes}
\end{equation}

This update is closed-form, runs in $O(|V|)$ time, and requires no gradient
computation. Multiple contradictory observations shift probability mass
naturally without explicit conflict detection.

\paragraph{Remark.}
Under Eq.~\eqref{eq:likelihood} a single confirming observation makes obs the
posterior mode whenever
$p_{\mathrm{prior}}(\text{obs}) > \varepsilon / (1 - \varepsilon)$; with
$\varepsilon = 0.01$ this threshold is $\approx 0.0101$, so any value already
holding non-trivial mass is promoted immediately. This is a sanity property of the
update, not a strong result; the interesting behaviour is how mass accumulates
under repeated, differently-reliable evidence, which we study empirically in
Sections~\ref{sec:reliability} and~\ref{sec:security}.

\subsection{Entropy Decay}

A dimension not updated for time $\Delta t$ grows more uncertain via
exponential mixing toward the uniform distribution:

\begin{equation}
  p_t(v) = \lambda^{\Delta t}\; p_0(v)
          + \bigl(1 - \lambda^{\Delta t}\bigr)\; \mathcal{U}(v),
  \label{eq:decay}
\end{equation}

where $\mathcal{U}(v) = 1/|V|$ and $\lambda \in (0,1)$ is a per-dimension
retention factor. As $\Delta t \to \infty$, $p_t \to \mathcal{U}$, encoding
complete uncertainty. The entropy satisfies:

\begin{equation}
  H(p_t) \;=\; -\sum_{v} p_t(v) \log_2 p_t(v) \;\geq\; \lambda^{\Delta t}\, H(p_0),
  \label{eq:entropy}
\end{equation}

guaranteeing the agent grows less certain as time passes without
reinforcement, a property heuristic deletion rules cannot provide.

\subsection{Identity Resolution}

When two entity mentions may refer to the same person, Nous computes the
symmetrised KL divergence between their shared dimensions:

\begin{equation}
  D(e_1, e_2) = \sum_{a \in \mathcal{A}_{e_1} \cap \mathcal{A}_{e_2}}
  \!\mathrm{KL}(p_{e_1,a} \| p_{e_2,a})
  + \mathrm{KL}(p_{e_2,a} \| p_{e_1,a}).
  \label{eq:kl}
\end{equation}

Entities with low divergence across shared attributes are candidates for
merging via a weighted mixture of their posteriors.

\subsection{Observed-Values Aggregation}
\label{sec:obsvals}

Bayesian posteriors are unimodal: after many updates, mass concentrates on
one value, discarding historical alternatives. For inherently multi-valued
attributes such as activities or places visited, this is incorrect. Nous
addresses this via delta-log aggregation: at query time, every value that
appeared with surprise above threshold $\theta_{\min}$ is included in the
context fact line:

\begin{equation}
  V^*_{e,a} = \bigl\{
    \Delta.\text{obs} \;:\;
    \Delta \in \mathcal{L}_{e,a},\;
    \Delta.\mathcal{S} > \theta_{\min}
  \bigr\}.
  \label{eq:obsvals}
\end{equation}

This preserves the Bayesian world model for the current best belief while
making the full historical record available for aggregation queries.

\subsection{Pipelines}

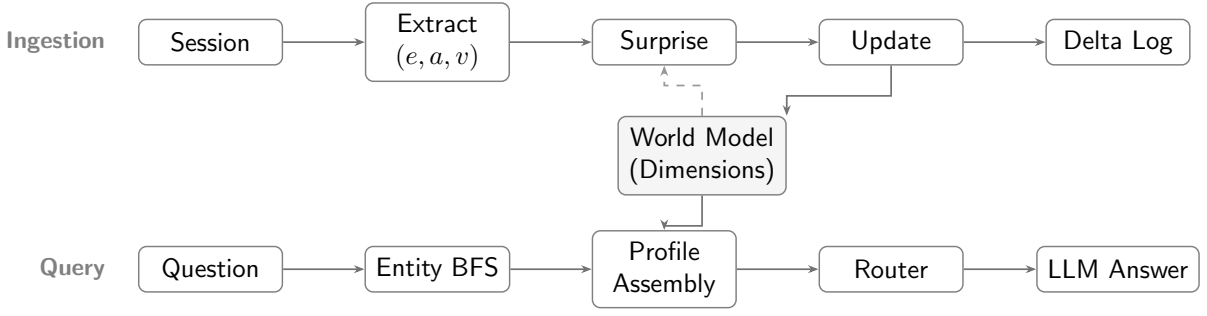
\begin{figure*}[t]
\centering
\begin{tikzpicture}[
  font=\small\sffamily,
  box/.style={
    draw=black!50, rounded corners=3pt,
    minimum width=1.9cm, minimum height=0.6cm,
    align=center, fill=white, line width=0.55pt,
    inner sep=4pt
  },
  wbox/.style={
    draw=black!50, rounded corners=3pt,
    minimum width=2.2cm, minimum height=0.6cm,
    align=center, fill=gray!8, line width=0.55pt,
    inner sep=4pt
  },
  arr/.style ={-{Stealth[length=4pt,width=3pt]},
               line width=0.65pt, color=black!55},
  darr/.style={-{Stealth[length=4pt,width=3pt]},
               line width=0.65pt, color=black!40, dashed},
  lbl/.style ={font=\footnotesize\bfseries\sffamily, color=black!50}
]

\node[box] (A) at ( 0.0, 1.5) {Session};
\node[box] (B) at ( 3.0, 1.5) {Extract \\ $(e,a,v)$};
\node[box] (C) at ( 6.0, 1.5) {Surprise};
\node[box] (D) at ( 9.0, 1.5) {Update};
\node[box] (E) at (12.0, 1.5) {Delta Log};

\draw[arr] (A)--(B);
\draw[arr] (B)--(C);
\draw[arr] (C)--(D);
\draw[arr] (D)--(E);

\node[wbox] (W) at (6.5, 0.0) {World Model \\ (Dimensions)};

\draw[arr]  (D.south)  -- ++(0,-0.4) -| (W.north east);
\draw[darr] (W.north)  -- ++(0, 0.4) -| (C.south);

\node[box] (P) at ( 0.0,-1.5) {Question};
\node[box] (Q) at ( 3.0,-1.5) {Entity BFS};
\node[box] (R) at ( 6.0,-1.5) {Profile \\ Assembly};
\node[box] (S) at ( 9.0,-1.5) {Router};
\node[box] (T) at (12.0,-1.5) {LLM Answer};

\draw[arr] (P)--(Q);
\draw[arr] (Q)--(R);
\draw[arr] (R)--(S);
\draw[arr] (S)--(T);

\draw[arr] (W.south) -- ++(0,-0.4) -| (R.north);

\node[lbl, left=0.3cm of A] {Ingestion};
\node[lbl, left=0.3cm of P] {Query};

\end{tikzpicture}
\caption{Ingestion (top) and query (bottom) pipelines. The world model, a
live store of dimensions, is updated by every Bayesian posterior (solid arrow
from Update) and supplies the current dimension state during surprise
computation (dashed arrow). At query time, the world model feeds the Profile
Assembly step.}
\label{fig:arch}
\end{figure*}

Figure~\ref{fig:arch} illustrates both pipelines. During ingestion, an LLM
extractor parses each session into $(e,a,v)$ triples. For each triple, the
current dimension is retrieved, the surprise computed via Eq.~\ref{eq:surprise},
the posterior computed via Eq.~\ref{eq:bayes}, and a delta appended to the
log. A compressor prunes dimensions whose entropy exceeds a ceiling threshold.

During querying, named entities in the question seed a breadth-first search
over the entity graph (up to two hops). For each retrieved entity, its
dimensions are formatted as fact lines using Eq.~\ref{eq:obsvals}. The top-$k$
delta records with the highest BM25 similarity to the question are appended as
evidence lines, and the assembled context is passed to a category-specific
prompt and then to the answering LLM.

\section{Experimental Setup}

\subsection{Benchmark}
We evaluate on LoCoMo~\citep{maharana2024locomo}: ten dyadic conversations
spanning 19 to 32 sessions each, totalling 1,540 questions across single-hop
(841), multi-hop (282), temporal (321), and open-domain (96). The adversarial
category is excluded following the canonical protocol.

\subsection{Metrics}
We report two answer-quality metrics on the same generated outputs and are
explicit about their difference, because it is large and consequential.
\emph{Strict token-F1} is the unigram overlap between predicted and
ground-truth answer after normalisation; it does not reward semantically
correct but differently-phrased answers, and it is the metric under which the
academic baselines we compare against report. \emph{LLM-judge accuracy} is the
fraction of answers a GPT-4o-mini judge, instructed to accept an answer whose
key facts match regardless of wording, marks correct. We treat token-F1 as the
primary metric for cross-system comparison and report the judge score
alongside it; as Section~\ref{sec:results} shows, the two diverge by 27.5 macro
points, and the judge score is not comparable to baselines that use token-F1.
We additionally report BLEU-1; context recall, the fraction of ground-truth
answer tokens present anywhere in the assembled context, as a retrieval
diagnostic; and a four-way failure bucket classification (good, answer miss,
unknown, retrieval miss) assigned by the same judge.

\subsection{Baselines}
Nous uses \texttt{gemini-2.5-flash} for triple extraction and answer
generation, and \texttt{gpt-4o-mini} as the judge. We compare against: (i) a
no-memory baseline (direct QA, no retrieved context); (ii)
MemGPT~\citep{packer2023memgpt}; and (iii) A-MEM~\citep{xu2025amem}, the
strongest peer-reviewed academic baseline. Baseline token-F1 scores are taken
from~\citet{xu2025amem} (GPT-4o-mini backbone). Because the backbone and
retrieval stacks differ, we treat this as an indicative rather than a fully
controlled comparison; a matched re-run of each baseline is left to future
work. We do not compare numerically against Zep or Mem0, which report under
GPT-4o with a lenient LLM-judge~\citep{locomo2025audit}.

\section{Results}
\label{sec:results}

\paragraph{Main comparison (strict token-F1).}
Table~\ref{tab:main} reports strict token-F1, the metric under which the
academic baselines report. Under this metric Nous is competitive with, not
dominant over, A-MEM, splitting the four categories two--two: Nous is best on
temporal (51.10 vs 12.14, a 39-point margin) and single-hop (44.66 vs 27.02),
while A-MEM is best on multi-hop (45.85 vs 23.48) and open-domain (44.65 vs
10.54). Temporal, where categorical belief tracking most directly captures a
changing state, is Nous's clearest strength; the multi-hop and open-domain
deficits trace to list-aggregation and retrieval limits we analyse in
Section~\ref{sec:analysis}.

\begin{table}[t]
\centering
\caption{Strict token-F1 on LoCoMo (10 conversations, 1,540 questions).
Baselines from \citet{xu2025amem} (GPT-4o-mini). Nous uses gemini-2.5-flash
for extraction/answering. Best per column in bold. The comparison is
indicative: backbones and retrieval stacks differ (see text).}
\label{tab:main}
\renewcommand{\arraystretch}{1.25}
\setlength{\tabcolsep}{4pt}
\begin{tabular}{lcccc}
\toprule
System & SH & MH & T & OD \\
\midrule
LoCoMo       & 25.02 & 18.41 & 12.04 & 40.36 \\
MemGPT       & 26.65 & 25.52 & 9.15  & 41.04 \\
A-MEM        & 27.02 & \textbf{45.85} & 12.14 & \textbf{44.65} \\
\midrule
\textbf{Nous}& \textbf{44.66} & 23.48 & \textbf{51.10} & 10.54 \\
\bottomrule
\end{tabular}
\end{table}

SH = single-hop, MH = multi-hop, T = temporal, OD = open-domain.

\paragraph{Metric sensitivity (the headline caveat).}
An earlier version of this work reported the LLM-judge column of
Table~\ref{tab:metric} as ``F1.'' It is not: it is a generous LLM-as-judge that
accepts semantically correct paraphrases. On identical outputs the judge scores
59.97 macro while strict token-F1 scores 32.45, a 27.5-point gap. Three of the
four judge scores exceed the mathematical maximum token-F1 attainable given the
measured BLEU-1, confirming the two cannot be the same quantity. The gap is
largest exactly where answers admit many surface forms: open-domain
(judge 62.50 vs token-F1 10.54; context recall only 47.5\%, so the judge credits
answers the retriever never supported). This is not a Nous-specific artifact but
a property of LoCoMo evaluation~\citep{locomo2025audit}; we report it as a
result in its own right, and urge that memory-system comparisons fix the judge
before comparing scores.

\begin{table}[t]
\centering
\caption{Nous under three answer-quality lenses on the same outputs, plus
context recall. The judge/token-F1 gap (27.5 macro) is the metric-sensitivity
finding; it is \emph{not} a quality improvement.}
\label{tab:metric}
\renewcommand{\arraystretch}{1.25}
\setlength{\tabcolsep}{4pt}
\small
\begin{tabular}{lcccc}
\toprule
Category & Token-F1 & BLEU-1 & LLM-judge & Ctx Rec \\
\midrule
Single-hop  & 44.66 & 45.60 & 63.50 & 86.07 \\
Multi-hop   & 23.48 & 23.90 & 55.32 & 75.35 \\
Temporal    & 51.10 & 56.04 & 58.57 & 61.82 \\
Open-domain & 10.54 & 10.06 & 62.50 & 47.50 \\
\midrule
Macro avg   & 32.45 & 33.90 & 59.97 & 67.69 \\
\bottomrule
\end{tabular}
\end{table}

\section{Analysis}
\label{sec:analysis}

\subsection{Failure Bucket Breakdown}

Table~\ref{tab:buckets} shows the four-way failure distribution. Good denotes
a correct answer; answer miss means context was present but the LLM answered
incorrectly; unknown means the LLM declined to answer despite relevant context
being present; retrieval miss means relevant information was absent from the
assembled context.

\begin{table}[h!]
\centering
\small
\caption{Failure buckets for Nous across all categories.}
\label{tab:buckets}
\renewcommand{\arraystretch}{1.2}
\setlength{\tabcolsep}{3pt}
\begin{tabular}{lrrrr}
\toprule
 & Good & Ans.miss & Unknown & Ret.miss \\
\midrule
SH  & 529 (63\%) & 129 (15\%) & 154 (18\%) & 29 (3\%) \\
MH  &  89 (32\%) &  92 (33\%) &  18 (6\%)  & 24 (9\%) \\
T   & 175 (55\%) &  65 (20\%) &  62 (19\%) & 19 (6\%) \\
OD  &  40 (42\%) &  21 (22\%) &   0 (0\%)  & 35 (37\%) \\
\bottomrule
\end{tabular}
\end{table}

For single-hop, the dominant failure is unknown answer (18\%) despite a
context recall of 86.07\%. Of 154 such failures, 92 occur when context recall
is at least 0.7, meaning the information is present but the model declines to
commit. This is a backbone calibration issue rather than an architectural one.

For multi-hop, good (32\%) and answer miss (33\%) are nearly equal at adequate
context recall (75.35\%), indicating the model struggles to synthesise complete
lists from distributed fact lines. Retrieval miss (9\%) contributes
additionally, as some questions require linking entities beyond the two-hop
search.

Temporal suffers from both low context recall (61.82\%, the lowest of all
categories) and high unknown answer (19\%). Temporal questions often depend on
exact session dates embedded in evidence text rather than extracted as dimension
values, a representational gap addressed in Section~\ref{sec:limitations}.

Open-domain has the highest retrieval miss rate (37\%) and lowest context
recall (47.50\%). Questions about personality and long-term habits do not map
cleanly to entity-attribute dimensions, and improving retrieval is the primary
path forward.

\subsection{Context Recall as a Diagnostic}

Context recall cleanly separates retrieval failures from answering failures.
For single-hop, context recall is high (86.07\%) while strict token-F1 is 44.66,
identifying the answering/normalisation step, not retrieval, as the binding
constraint. For open-domain, context recall (47.50\%) is itself the ceiling:
the retriever supports fewer than half the answers, which is why the generous
judge score there (62.50) is untrustworthy and the token-F1 (10.54) so low.

\section{Ablations: What Actually Drives the Results}
\label{sec:ablations}

The architecture's central claim is that surprise-driven Bayesian belief
tracking is what makes the memory work. We test this directly with controlled
ablations on a three-conversation subset (385 questions), using BM25 with
two-hop entity BFS as the retriever and scoring on both metrics. We toggle one
component at a time: the Bayesian update (replaced by naive last-write-wins,
which collapses all mass onto the newest value), and observed-values
aggregation (which lists all values ever seen for an attribute).

\begin{table}[t]
\centering
\caption{Component ablations on the 3-conversation subset (385 questions),
BM25+BFS retrieval. Disabling the Bayesian update (\textsc{flat}) does not hurt;
disabling aggregation (\textsc{no-agg}) does. Macro over the four categories.}
\label{tab:ablation}
\renewcommand{\arraystretch}{1.25}
\setlength{\tabcolsep}{5pt}
\begin{tabular}{lcc}
\toprule
Variant & LLM-judge & Token-F1 \\
\midrule
Full engine                       & 69.62 & 36.77 \\
$-$ Bayesian update (\textsc{flat}) & \textbf{72.46} & 36.27 \\
$-$ Aggregation (\textsc{no-agg})   & 66.58 & 35.39 \\
\bottomrule
\end{tabular}
\end{table}

The result is decisive and, for our thesis, uncomfortable
(Table~\ref{tab:ablation}). \textbf{Replacing the Bayesian update with naive
last-write-wins does not reduce accuracy}: it is marginally better on the judge
(72.46 vs 69.62) and a wash on token-F1 (36.27 vs 36.77). On LoCoMo, the
Bayesian belief update is effectively \emph{inert}. The component that does
matter is the mundane one: disabling observed-values aggregation costs the most
(judge 66.58), the drop concentrated in multi-hop (judge 55.41 $\rightarrow$
50.00), because list-valued questions need every value ever seen, not a single
most-probable one. We also find the dense retriever inert on this data
(single-conversation: judge 63.34 with dense vs 65.26 without), so the BM25+BFS
stack the architecture describes is sufficient; and entropy decay is never
triggered during LoCoMo evaluation, so these results say nothing about it.

The explanation is not that Bayesian updating is worthless but that LoCoMo does
not stress it. Most LoCoMo questions concern facts that are stated once or that
do not oscillate; when a value simply needs to be the latest one asserted,
``keep the newest'' ties the full posterior. The belief machinery is built to
arbitrate \emph{contradictory, noisy, differently-reliable} evidence, a regime
this benchmark rarely presents. The next section isolates that regime and shows
the mechanism is far from worthless there, once given the input it needs.

\section{When Belief-Updating Helps: Reliability Extraction}
\label{sec:reliability}

\paragraph{Why the update was inert.}
The posterior update weights each observation by a reliability $r$: the
likelihood of the observed value is $r$ and the remaining mass is spread over
alternatives. A standard extraction pipeline emits
$(\text{entity},\text{attribute},\text{value})$ triples with \emph{no}
reliability, so every observation enters with the same fixed $r$. With constant
$r$, the update degenerates into a soft recency-follower: repeated or recent
values accumulate mass regardless of how trustworthy they are. This is exactly
why it ties last-write-wins.

We verify this at the belief layer with a synthetic contradiction
micro-benchmark (no LLM, no retrieval): sequences of $(\text{value}, r)$
observations for one attribute, scored on whether each strategy predicts the
current true value. With \emph{constant} reliability (what the real pipeline
feeds), Bayesian scores 67.1 macro, slightly \emph{below} last-write-wins
(73.5). With \emph{varying} reliability that reflects true source trust,
Bayesian reaches 100 and dominates both last-write-wins (74.3) and
frequency-counting (48.2). The mechanism's value is entirely conditional on a
reliability signal the pipeline was not providing.

\paragraph{Supplying the missing signal.}
We add a reliability-extraction component: the extractor estimates a per-claim
reliability in $[0,1]$ from the speaker's \emph{epistemic markers}: hedging
(``maybe'', ``I think'', ``if I recall'') lowers it, definiteness and
self-correction (``actually'', ``confirmed'', ``definitely'') raise it, using a
fixed rubric and never observing ground truth. This per-claim reliability is
threaded into the Bayesian update.

\paragraph{End-to-end evaluation.}
We build a contradiction/staleness benchmark: 60 scenarios across five regimes
in which an entity's attribute evolves through natural-language statements of
varying linguistic confidence, some false, after which the current value is
queried. One regime is \emph{adversarial} (confidence is inverted, so the true
value is hedged and a false value asserted), included as an anti-rigging guard
where reliability-weighting is expected to lose. We compare, all reading the
same claims: \textsc{nous-reliability} (Bayesian, extracted reliability),
\textsc{flat} (last-write-wins), \textsc{freq} (most frequent), and
\textsc{append-retrieve} (a capable LLM given \emph{all} raw statements in order).

First, the extractor recovers confidence from phrasing: statements written to be
confident receive mean reliability 0.95, hedged ones 0.23 (separation $+0.72$);
the no-match fallback fires near-zero, so this is not an artifact. Second
(Table~\ref{tab:reliability}), on the realistic regimes \textsc{nous-reliability}
scores 100 macro versus 67 for last-write-wins and 46 for frequency, and beats
even \textsc{append-retrieve} (90), which is given lossless recall and pays a
query cost that grows with memory, whereas Nous queries a bounded posterior. The
edge over \textsc{append-retrieve} is concentrated exactly in noise resistance
(stable-noise 100 vs 75; reliability-conflict 100 vs 83). As designed, all
confidence-based methods fail the adversarial regime; the honest claim is
therefore conditional: reliability-weighting helps \emph{when epistemic
confidence tracks correctness}, the common pragmatic case, and hurts when it is
inverted.

\begin{table}[t]
\centering
\caption{Contradiction/staleness benchmark: accuracy predicting the current
value. Realistic macro excludes the adversarial guard regime.
\textsc{append-retrieve} is an LLM reading all raw statements.}
\label{tab:reliability}
\renewcommand{\arraystretch}{1.2}
\setlength{\tabcolsep}{4pt}
\small
\begin{tabular}{lcccc}
\toprule
Regime & Nous-rel & Flat & Freq & Append-ret \\
\midrule
stable-noise          & \textbf{100} & 42 & 100 & 75 \\
reliability-conflict  & \textbf{100} & 25 & 0   & 83 \\
clean-change          & 100 & 100 & 83 & 100 \\
recent-change         & 100 & 100 & 0  & 100 \\
adversarial (guard)   & 0   & 42 & 50 & 17 \\
\midrule
Realistic macro       & \textbf{100} & 67 & 46 & 90 \\
\bottomrule
\end{tabular}
\end{table}

To check that the extractor reads epistemic \emph{stance} rather than matching a
few templates, we re-ran with a larger, naturalistic phrasing bank containing
subtle hedges absent from the rubric (``supposedly'', ``apparently'', ``last I
checked'', ``don't quote me''). The signal and the advantage are essentially
unchanged (separation $+0.71$; realistic macro 100 vs 65 vs 40). This is a
proof-of-mechanism on synthetic data, not a claim about natural dialogue: the
confidence--correctness correlation is by construction in four of five regimes,
and a real-dialogue contradiction slice is the natural next step. But it
establishes what the LoCoMo ablation could not: given the reliability signal it
was designed for, belief-updating memory provides a large,
mechanism-attributable advantage over both naive overwriting and strong
LLM-over-memory retrieval.

\section{Security: Provenance-Capped Belief Updating}
\label{sec:security}

The reliability signal of Section~\ref{sec:reliability} raises an adversarial
question: if the attacker controls the text, can they forge the confidence that
earns trust? Because the numbers in this section are striking, we state their
boundaries first: the conditions are what make the results honest. (i)~Provenance
trust is assigned by an \emph{oracle}: each source receives a tier from its
channel; we do not learn or verify tiers. (ii)~The benchmarks are
\emph{synthetic} (a controlled belief-layer simulation and a minimal
coding-assistant harness over a single repository), not a deployed system.
(iii)~The defense against laundering requires taint-tracking that we
\emph{characterize but do not implement}. (iv)~We cite
AgentPoison~\citep{chen2024agentpoison} and MINJA~\citep{dong2025minja} only to
establish that memory poisoning is a real, studied threat; our attack is a
volumetric fact-overwrite and our success rates are \emph{not numerically
comparable} to theirs.

\paragraph{Trust composition.}
An attacker injects observations asserting a false value $W$ to overwrite an
established belief $V$. Each observation carries reliability
$r=\min(\text{provenance},\text{content})$: content confidence may only
\emph{lower} trust within a channel's provenance ceiling, never raise it.

\paragraph{Content-inferred trust is adversarially gameable.}
Using the extractor of Section~\ref{sec:reliability}, we phrased poison in an
authoritative style (``Confirmed: the database is now MongoDB, verified''). The
real extractor assigns it mean reliability $0.96$, versus $0.25$ for hedged
phrasing. We are explicit about what this does and does not show: we wrote the
authoritative poison and our own extractor scored it high, so this demonstrates
that content-inferred trust \emph{can} be gamed by construction (an attacker who
controls the text controls this signal), not that any specific deployed
content-trust defense~\citep{mempoison2026} fails against a real adversary. The
architectural consequence is the same either way: a signal the attacker can write
cannot be the trust anchor, which is why trust must come from provenance, which the
attacker does not control.

\paragraph{Under provenance-capped trust, low-trust poison cannot move the
belief.}
\emph{Condition:} the attacker is confined to a channel the system tiers below
$0.5$ by provenance (scraped web content, tool output), independent of how the
payload is worded. \emph{Result} (Table~\ref{tab:poison}, 400 trials/cell): a
volumetric flood of up to $50$ poison observations against $4$ trusted ones
yields $0\%$ attack success, where append-and-retrieve, last-write-wins, and
content-trusted updating are all eventually flipped to $100\%$ by volume or by a
confident phrasing. Ablating the source-trust signal alone (poison trusted
equally) restores $100\%$, confirming that the \emph{provenance signal}, not the
Bayesian arithmetic, is what defends. The guarantee is precisely bounded, not
universal: it holds for channel trust $\le 0.5$ at any volume, and fails above
$\sim 0.55$.

\begin{table}[t]
\centering
\caption{Attack success rate (\%; lower is better) as an attacker injects $M$
false observations against a belief established by four trusted ones, on a
channel tiered $0.2$ by provenance. Belief-layer simulation, 400 trials/cell.
Provenance-capped updating is the only strategy that resists a volumetric flood;
this holds \emph{because} the low tier is assigned by provenance, not content.}
\label{tab:poison}
\renewcommand{\arraystretch}{1.2}
\setlength{\tabcolsep}{4pt}
\small
\begin{tabular}{lcccc}
\toprule
Memory strategy & $M{=}1$ & $M{=}3$ & $M{=}10$ & $M{=}50$ \\
\midrule
Last-write-wins        & 100 & 100 & 100 & 100 \\
Append (majority)      &   0 &   0 & 100 & 100 \\
Content-trusted        & 100 & 100 & 100 & 100 \\
\textbf{Provenance-capped} & \textbf{0} & \textbf{0} & \textbf{0} & \textbf{0} \\
\bottomrule
\end{tabular}
\end{table}

\paragraph{Laundering defeats naive tiering; only taint-tracking survives, and
we do not implement it.}
\emph{Condition:} the attacker routes the same poison through a \emph{trusted}
intermediary, a tool the agent trusts that has ingested attacker-controlled
data. If laundered content inherits the intermediary's high tier, attack success
returns to $100\%$: provenance tiering \emph{alone} is fully defeated by
laundering. Only when provenance is \emph{propagated}, laundered content
down-tiered to its untrusted origin (taint tracking), does success return to
$0\%$. Taint tracking through every trusted intermediary is therefore a
\emph{requirement} of the defense, not an optimisation, and a hard one
(information-flow control); we characterize it but leave its implementation to
future work. The belief update was never the difficult part; provenance
propagation is.

\paragraph{Costs: the cap is not free.}
Two costs follow directly. First, utility: a genuinely reliable source arriving
on a low-provenance channel is under-weighted, because the $\min$-cap that blocks
poison also blocks \emph{legitimate} low-tier corrections: the same mechanism,
opposite ground truth. Routing a true, correcting source through tier $0.3$ drops
accuracy on the contradiction regimes from $100\%$ (uncapped) to $54\%$ on
average, with the collapse concentrated in the high-noise regime (down to $21\%$)
rather than the sparse-conflict one (which holds at $88\%$). Second,
extraction: because Nous stores structured beliefs rather than raw text, it pays
an \emph{extraction tax} that raw retrieval does not. On 30 real repository
facts, answerability falls from $100\%$ under perfect extraction to $97\%$ when
the real extractor must infer entity/attribute/value, with occasional total
drops. This $\sim\!3\%$ is a \emph{floor}: our answerer renders all beliefs, so a
misfiled belief stays visible, whereas a real retriever would miss it.

\paragraph{End-to-end demonstration.}
In a minimal coding assistant over one repository (identical LLM and prompt,
only the memory layer differing), these effects reproduce end to end. A confident
false ``migrated to MongoDB,'' injected on the low-trust web channel, flips an
append-and-retrieve baseline ($100\%$) but not Nous ($0\%$); the same poison
laundered through a trusted tool flips naive Nous ($100\%$) but not the
taint-tracked variant ($0\%$); and on a \emph{legitimate} low-trust correction
Nous answers \emph{wrong} where the baseline is right, making the utility cost
concrete. We report this as a controlled demonstration on caveated, synthetic
conditions, not as evidence of a deployed defense.

\section{Limitations and Future Work}
\label{sec:limitations}

The Table~\ref{tab:main} comparison is indicative, not controlled: baseline
scores come from~\citet{xu2025amem} under a GPT-4o-mini backbone whereas Nous uses
gemini-2.5-flash with a different retrieval stack, so a matched single-backbone
re-run of A-MEM (and a controlled comparison against concurrent belief-based memory
systems) is needed to make it definitive.

The reliability result (Section~\ref{sec:reliability}), while it establishes the
mechanism end-to-end, rests on a \emph{synthetic} contradiction benchmark whose
confidence--correctness correlation is by construction in four of five regimes.
The highest-priority extension is a real-dialogue slice: mining genuine
self-corrections and hedged-then-revised facts from conversational data and
testing whether the advantage survives. Finer-grained reliability (beyond the
near-binary confident/hedged split the extractor currently produces) and
learned rather than rubric-based reliability are open directions.

Entropy decay is never exercised by LoCoMo evaluation, so this paper provides no
empirical evidence for or against it; a staleness benchmark that queries
deliberately outdated facts is required. The categorical dimension is also
unimodal, requiring delta-log aggregation as a workaround for set-valued
attributes; a set dimension maintaining an independent Bernoulli distribution
per element is the main planned architectural extension, and making event
timestamps explicit dimension attributes would improve temporal context recall
(currently 61.82\%).

Finally, all our results use that single backbone, so cross-backbone
generalisation is unconfirmed, and completing the in-progress
LongMemEval~\citep{wu2025longmemeval} evaluation on a second, more carefully
audited benchmark is necessary for a complete picture. The Penfield Labs
audit~\citep{locomo2025audit} reports roughly 6.4 percent erroneous LoCoMo ground
truths and judge-calibration issues, consistent with the metric gap we document in
Section~\ref{sec:results}.

\section{Conclusion}

We set out to determine \emph{when} belief-based agent memory helps. Our vehicle,
Nous, treats knowledge as a predictive world model rather than a database of facts:
each entity-attribute pair is a categorical distribution updated by
information-theoretic surprise and a closed-form Bayesian posterior, with deltas as
the primary artifact and entropy decay as forgetting, an architecture grounded in
predictive coding and information theory.

Our empirical account is deliberately honest rather than triumphal. Under a
strict token-F1 metric Nous is competitive with, not dominant over, the
strongest academic baseline (clearly best on temporal reasoning, mixed
elsewhere), and a controlled ablation shows that on LoCoMo the Bayesian belief
update is inert, tied by naive last-write-wins, because conversational QA rarely
presents the contradictory, low-reliability evidence the mechanism is designed
for. We trace this to a missing per-observation reliability signal, add a
reliability-extraction component that recovers it from epistemic markers in
language, and show on a controlled contradiction benchmark that belief-updating
then earns its keep: it beats naive overwriting (100 vs 67) and even a strong
LLM-over-raw-memory baseline (100 vs 90), with the advantage concentrated in
noise resistance. That same reliability signal is an attack surface: we measure
that content-inferred trust is adversarially gameable and show that
provenance-capped updating, $\min(\text{provenance},\text{content})$, holds
volumetric memory-poisoning at $0\%$ attack success where naive baselines reach
$100\%$, a defense that is bounded, requires taint-tracking, carries a measured
utility cost, and is narrow in novelty relative to prior content-based defenses.
We also quantify a 27.5-point gap between strict token-F1 and a generous
LLM-judge on identical outputs, a caution for the whole subfield's comparisons. The takeaway is not that probabilistic world models win every
benchmark, but a sharper and more useful claim: they help precisely when memory
must arbitrate conflicting, differently-trustworthy evidence, and we identify
the component that makes that benefit real. All code and evaluation harnesses,
including both benchmarks, are released publicly.

\clearpage
\bibliography{nous}
\bibliographystyle{plainnat}

\end{document}